\documentclass[manuscript,screen,authorversion,nonacm]{acmart}
\usepackage{subfig}
\AtBeginDocument{%
  \providecommand\BibTeX{{%
    \normalfont B\kern-0.5em{\scshape i\kern-0.25em b}\kern-0.8em\TeX}}}

\setcopyright{acmcopyright}
\copyrightyear{2018}
\acmYear{2021}

\acmConference[IUI '21]{IUI '21: ACM Intelligent User Interfaces}
\acmPrice{15.00}
\acmISBN{978-1-4503-XXXX-X/18/06}

\begin{document}

\title{FakeBuster: A DeepFakes Detection Tool for Video Conferencing Scenarios}






\author{Vineet Mehta}
\email{2016csb1063@iitrpr.ac.in}
\affiliation{%
  \institution{Indian Institute of Technology Ropar}
  \city{Ropar}
  \country{India}
 }

\author{Parul Gupta}
\affiliation{%
  \institution{Indian Institute of Technology Ropar}
  \country{India}
 }
\email{2016csb1048@iitrpr.ac.in}

\author{Ramanathan Subramanian}
\affiliation{%
 \institution{Indian Institute of Techhnology Ropar}
 \country{India}
}

\author{Abhinav Dhall}
\affiliation{%
  \institution{Monash University}
  \country{Australia}
  }
\affiliation{%
  \institution{Indian Institute of Technology Ropar}
  \country{India}
  }


\begin{abstract}
This paper proposes a new DeepFake detector \emph{FakeBuster} for detecting impostors during video conferencing and manipulated faces on social media. \emph{FakeBuster} is a standalone deep learning based solution, which enables a user to detect if another person's video is manipulated or spoofed during a video conferencing based meeting. This tool is independent of video conferencing solutions and has been tested with Zoom and Skype applications. It uses a 3D convolutional neural network for predicting video segment-wise fakeness scores. The network is trained on a combination of datasets such as Deeperforensics, DFDC, VoxCeleb, and deepfake videos created using locally captured (for video conferencing scenarios) images. This leads to different environments and perturbations in the dataset, which improves the generalization of the deepfake network.
\end{abstract}


\maketitle
\section{Introduction and Background}
Sophisticated artificial intelligence techniques have led to readily available manipulated media content which keeps evolving and is becoming increasingly difficult to be recognised. While their usage in spreading fake news, pornography and other such online content has been widely observed with major repercussions \cite{mstool}, recently, they have also found their way into video calling platforms through spoofing tools based on facial performance transfer \cite{siarohin2019first}. It enable individuals to mimic other persons through the real-time transfer of their expressions on the target person's frontal face image and the resulting deepfakes are quite convincing to the human eye. This may have serious implications, especially in the current situation of pandemic where video calling is the quintessential mode of personal as well as professional communication.

\begin{figure*}[!t]
\centering
    \includegraphics[width=\textwidth]{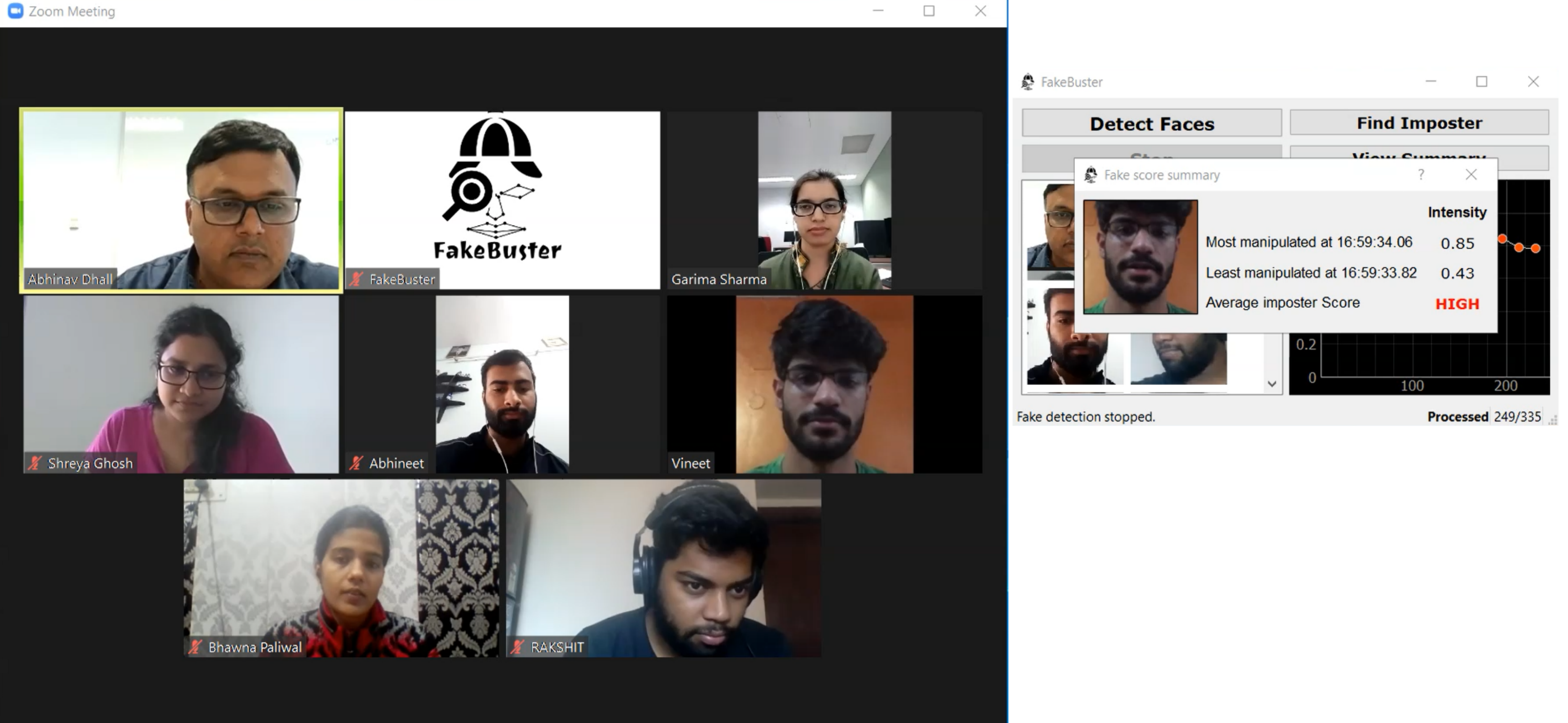}
    \caption[]{\emph{FakeBuster} tool enables detection of an impostor in online meetings. Here a user's video feed in a Zoom meeting is detected as an impostor by \emph{FakeBuster}.
    Please see the video demo of \emph{FakeBuster} on youtube\footnotemark. }\vspace{-4mm}
    \label{usecase}
\end{figure*}

There are few deepfake detection softwares such as the recently introduced Microsoft's Video Authenticator \cite{rdtool}. However, these are useful only for verifying the genuineness of pre-recorded videos. Also, Microsoft’s tool is not available for public use to prevent its misuse. To overcome these limitations, we present \emph{FakeBuster} - a deepfakes detection tool, which works in both offline (for a pre-existing video) as well as online scenarios (during video conferencing). A snapshot of the tool being used during a Zoom meeting is shown in Figure \ref{usecase}.

\textbf{Use Cases:} Tools for detection of deepfakes can help in identification of an impostor during events such as online examinations, video based authentication and job interviews. Organisations can use deepfakes detection tools to ensure the legitimacy of a candidate. The same tool can also be used to validate any media content seen online by the user on social media platforms such as YouTube.com and Twitter.com.

\section{FakeBuster}
\emph{FakeBuster} has a compact interface designed for ease of use alongside video calling applications and internet browsers. It has been developed using PyQt toolkit \cite{pyqt} which offers the following advantages:- a) It is a python binding of QT \cite{qt}, which is a cross-platform application development framework, and b) use of python language enables flexible integration of deep learning models. We have used the python MSS \cite{python-mss} library for screen recording, OpenCV \cite{opencv_library} for image processing, and Pytorch \cite{NEURIPS2019_9015} to train and test the deep learning models. The standalone aspect of the tool enables its usage with different video conferencing tools such as Zoom, Skype and Webex etc.

\footnotetext{Youtube link-\url{https://youtu.be/XZvybwXpm_g}}

\subsection{Workflow}
\begin{figure*}[!t]
\centering
    \subfloat[Step 1 - Face Select]{\includegraphics[width = .4\textwidth]{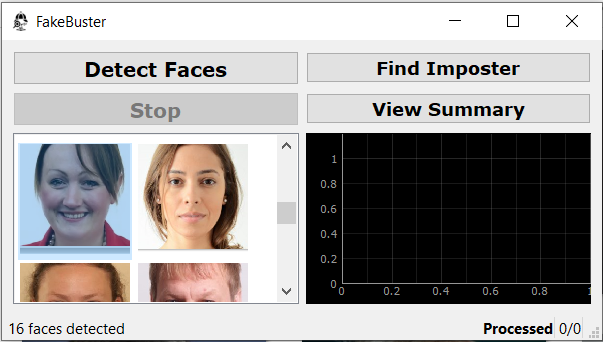}}
    \hskip .5em
    \subfloat[Step 2 - Find Imposter]{\includegraphics[width = .4\textwidth]{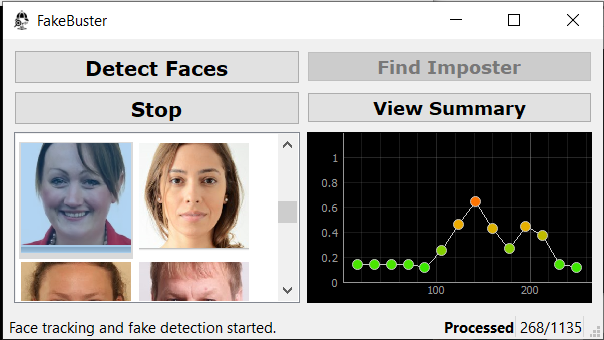}}
    \hskip.5em
    \subfloat[Step 3 - Analysis]{\includegraphics[width = .4\textwidth]{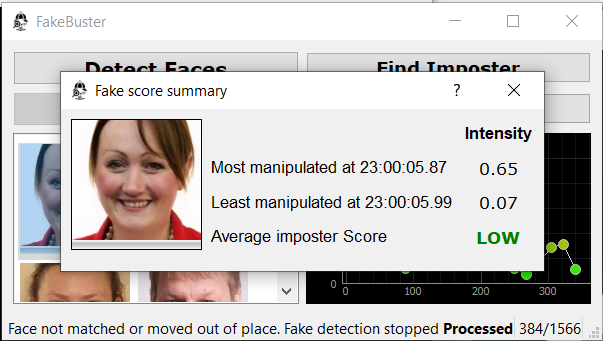}} \vspace{-1mm}
\caption{The workflow of \emph{FakeBuster}- a) User selects the faces detected by the tool from videoconferencing tool; b) User clicks on find impostor and the tool does prediction at snippet level; and c) User views the meta-analysis of fakeness of the video feed.}\label{workflow} \vspace{-3mm}
\end{figure*}

The tool works in three steps (Figure \ref{workflow}): a) The user clicks on "Detect Faces" and the tool detects and shows all the faces present in the screen. In case of change in the video conference software's view or position, the faces can be detected again by clicking on the "Detect Faces" button. The user then selects a face icon belonging to a person, whose video needs to checked; b) The user next clicks on the "Find Imposter" button to start the deep learning algorithm inference. The tool runs face capturing, segmentation, and deepfake prediction for each segment (set of frames) in the background. The time-series graph on the right side of the tool (see Fig. \ref{workflow} (b)) shows the prediction score at regular intervals. The prediction score varies between the range 0 to 1, which is color-coded, depicting the extent of face manipulation at a particular instant:- \emph{green} (no manipulation), \emph{yellow, orange}, and \emph{red} (high probability of presence of manipulation); c) The user can see the overall summary by clicking on the "View Summary" button. It opens a new dialog screen (see Fig. \ref{workflow} (c)), which shows the \emph{average imposter score, highest} and \emph{lowest manipulation intensity}, along with the \emph{occurrence time stamps}. The user can use this information to further take an action during the video conferencing meeting.

In the background, the imposter detection is started on a particular face, screen capture around the face position is computed. Further, \emph{FakeBuster} performs face tracking, face frames segmentation, and deepfake prediction on each video segment in the background. The prediction scores are aggregated, and the time-series graph is updated at the same time.

\subsection{Deepfake Detection}
\begin{figure*}[!t]
  \centering
  \includegraphics[width=80mm]{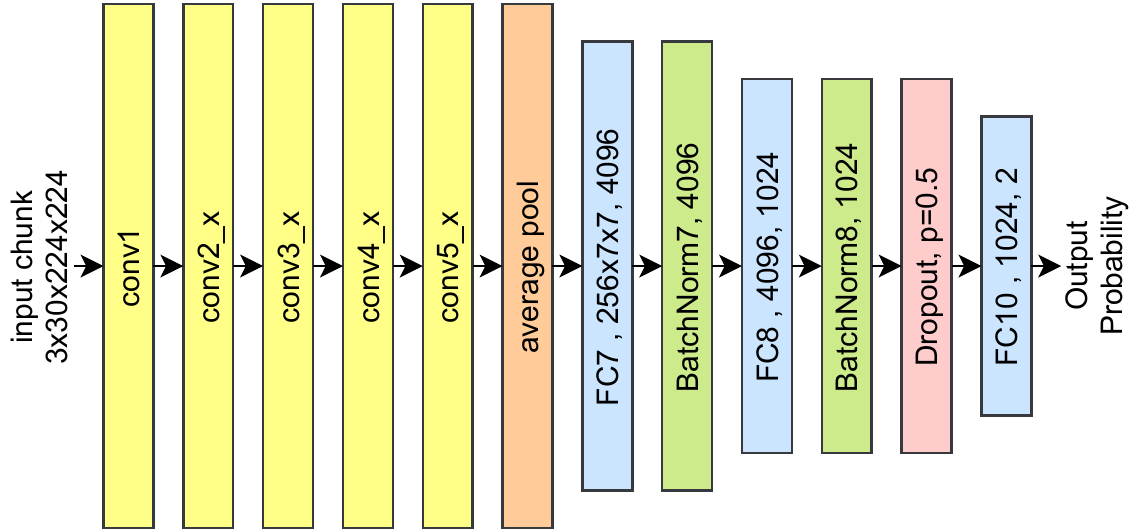}
  \caption{Deepfake detection network architecture.} \vspace{-3mm}
  \label{network}
\end{figure*}

For deepfake prediction on a given segment, we train a deep learning model i.e. 3D ResNet \cite{DBLP:journals/corr/abs-1711-09577} based visual stream architecture (see Fig. \ref{network}) followed by the recent state-of-the-art deepfake detection method by Chugh et al. \cite{10.1145/3394171.3413700}. The input to the network is a segment of 30 face frames and the output is the probability of the input chunk being fake.

There are various large scale deepfake datasets such as DFDC \cite{dolhansky2019deepfake} and Deeperforensics \cite{jiang2020deeperforensics10} consisting of fake videos generated using face-swapping techniques that generate good quality deepfakes but are never found to be used in online fashion. However, one of the successful online deepfake creator tools- Avatarify \cite{avatarify} performs face swapping in video conferencing apps using First Order Motion Model (FOMM)  proposed by Siarohin et al. \cite{siarohin2019first}. These online generated deepfake image content are relatively low in resolution. Moreover, the video conferencing scenario adds up different types of perturbations due to diversity in camera quality, network bandwidth, and artifacts by video encoding-decoding algorithms. Therefore, we created a separate dataset consisting of 10,000 real and 10,000 fake videos.

\textbf{Dataset:}
To generate facial performance transfer videos we used FOMM \cite{siarohin2019first} to swap faces. At first, we manually selected (to ensure frontal face position) 100 face videos from the pre-processed VoxCeleb \cite{Nagrani_2017} dataset.  We collected 25 images from 25 subjects, which are captured using the mobile cameras or laptop webcams to mimic the low image quality and pose as found during video calls. Moreover, we also selected 75 front-facing high-quality face images with plain background from Flickr-Faces-HQ Dataset (FFHQ) \cite{karras2019stylebased}. By swapping every image's face with the one in every video, we generated 10,000 fake videos.
To get an equal number of real videos, we selected 4105 face videos from the pre-processed VoxCeleb \cite{Nagrani_2017} dataset. The remaining 5895 face videos are obtained by processing front camera pose videos chosen from the Deeperforensics \cite{jiang2020deeperforensics10} dataset.

\textbf{Training and Results:}
We created a subject-independent 3:1 train/test split from the created dataset. For a better generalization of the network, we also used and performed a 3:1 train/test split of a balanced subset from the DFDC \cite{dolhansky2019deepfake} dataset consisting of distinct 10,000 fake and corresponding real video pairs. We mixed both of the training sets and trained the model for 20 epochs with a batch size of 8 and a learning rate of 0.001 along with Adam \cite{kingma2014adam} optimizer. The Area Under the ROC Curve for this model is 90.61 on the test set.

\subsection{Ethical Use and Privacy}
\emph{FakeBuster} performs screen recording, which provides feasibility to test any face video that appears on the screen. However, screen recording may breach the privacy of the subjects present in a video conferencing meeting. Therefore, it is important that the members of the meeting are made aware beforehand that such tools will be used. \emph{FakeBuster} is designed in such a way that it does not store any image or video in the file system. Moreover, the captured face images are buffered and removed immediately as the deepfake prediction are made on them.

\section{Conclusion and Future works}
To the best of our knowledge, \emph{FakeBuster} is one of the first tools for detecting impostors during video conferencing using deepfake detection technology. It is noted that the standalone tool works well with different video conferencing solutions such as Zoom and Skype. The tool can also be used for checking if an online video is fake or not in close to real-time. The real-time aspect is dependent on the GPU capability of the machine on which  \emph{FakeBuster} is installed. 

At present, \emph{FakeBuster} can evaluate only one face at a time, as part of the future work, multiple face processing will be implemented. This will help in reducing the user effort in validation of multiple faces. From deep learning perspective, we aim to make the network smaller. \emph{FakeBuster} uses 3D convolutions, which are computationally expensive. Another important direction is adding more data to the training process such that the network can generalise better to unseen scenarios. \emph{FakeBuster} analysis video only, we will experiment with multimodal deep learning based deepfakes detection networks \cite{10.1145/3394171.3413700} by adding audio information.


\end{document}